\pdfoutput=1 
\documentclass[letterpaper, 10 pt, conference]{ieeeconf}  % Comment this line out if you need a4paper

\IEEEoverridecommandlockouts                              % This command is only needed if 
\usepackage{amssymb}
\usepackage{amsmath}
\usepackage{bm}
\usepackage{svg}
\usepackage{hyperref}
\usepackage{cite}
\usepackage[normalem]{ulem}

\title{\LARGE \bf
Object-centric Representations for Interactive Online Learning with Non-Parametric Methods 
}

\author{Nikhil U. Shinde, Jacob Johnson, Sylvia Herbert, and Michael C. Yip% <-this % stops a space
\thanks{This work was supported by NDSEG and NSF award \#2045803. The authors are with the University of California, San Diego. \{\href{mailto:nshinde@eng.ucsd.edu}{nshinde}, \href{mailto:jjj025@eng.ucsd.edu}{jjj025},  \href{mailto:sherbert@ucsd.edu}{sherbert}, \href{mailto:yip@eng.ucsd.edu}{yip}\}@ucsd.edu.}}
\begin{document}

\maketitle
\thispagestyle{empty}
\pagestyle{empty}

%%%%%%%%%%%%%%%%%%%%%%%%%%%%%%%%%%%%%%%%%%%%%%%%%%%%%%%%%%%%%%%%%%%%%%%%%%%%%%%%
\begin{abstract}
% \begin{itemize}
%     \item Large offline learning-based models enable applications where robots succeed in interacting with objects to perform a task. However, these models tend to rely on fairly consistent and structured environments.
%     \item For more unstructured environments, an online learning component is necessary to gather and estimate information about objects in the environment in order to successfully interact with them.
%     \item However, online learning methods like Bayesian non-parametric models struggle with a changing environment (e.g. when an object is moved through an interaction)
%     \item We propose a composable object-centric representation for online learning. 
%     \item Allows for scalable online learning of interaction effects
%     \item We demonstrate on blabla some results blabla
% \end{itemize}
Large offline learning-based models have enabled robots to successfully interact with objects for a wide variety of tasks. 
However, these models rely on fairly consistent structured environments. 
For more unstructured environments, an online learning component is necessary to gather and estimate information about objects in the environment in order to successfully interact with them. 
Unfortunately, online learning methods like Bayesian non-parametric models struggle with changes in the environment, which is often the desired outcome of interaction-based tasks.
We propose using an object-centric representation for interactive online learning.
%, which enables scalable online learning of interaction effects. 
This representation is generated by transforming the robot’s actions into the object’s coordinate frame. 
We demonstrate how switching to this task-relevant space improves our ability to reason with the training data collected online, enabling scalable online learning of robot-object interactions. 
%\nsnote{Sylvia Please look at this: }
We showcase our method by successfully navigating a manipulator arm through an environment with multiple unknown objects without violating interaction-based constraints. 

\end{abstract}

%%%%%%%%%%%%%%%%%%%%%%%%%%%%%%%%%%%%%%%%%%%%%%%%%%%%%%%%%%%%%%%%%%%%%%%%%%%%%%%%
\section{Introduction}
% \shmargin{}{in the title it would be nice to have something in there about interaction}
% 1. Robot manipulation is an important field: 
% automated robot manipulation is already being adopted throughout industry to improve efficiency and accuracy across several manufacturing tasks. 
% These tasks vary from putting together automobiles in assembly lines to painting and welding. 
% \shnote{something about working well in complex, interactive tasks because of strong structure in the environment (e.g. assembly line). This enables methods like offline learning to work well.}

Automated robot manipulation is rapidly being adopted throughout industry to improve efficiency and accuracy across several manufacturing tasks \cite{BROGARDH200769,gupta2009industrial}.
For applications that require interaction with objects in the environment (e.g. assembling automobile components), current successful methods require highly structured and consistent environments, such as assembly lines.
%These tasks vary from putting together automobiles in assembly lines to painting and welding. 
%In particular, automation involving interaction works best in industrial tasks like assembly lines for known, rigid components where a strong structure can be imposed on the environment configuration and the objects within it. 
This structure enables established planning algorithms and offline learning-based models to work well. 
%2. Current push in robot manipulation. 
%Many recent works focus on using large offline learned models to improve robot automation. 
% - These works often give quite good performance. 
%They serve as a good base to understand and interact with the environment. 
Unfortunately, these offline methods are usually parametric and have the drawback of being nearly impossible to update online as new data is observed. 
In many unstructured environments that are not perfectly modeled, this becomes a drawback, as there are many attributes of the environment that are difficult to learn without actively interacting with the environment. 

% - \shnote{Navigating and interacting with objects in less structured environments like warehouses or construction sites much more challenging.}
Navigating and interacting with objects in less structured environments like warehouses, construction sites, or even a common household remains challenging. 
As an example, picture a household pantry with many opaque containers. 
%Without manipulating these containers it may be hard to get a sense of how they will respond to robot interaction. 
Multiple parameters (e.g. center of mass, friction coefficients) are difficult to estimate without active interaction, and may drastically affect the results of the interaction. 
In these cases, it becomes important to have an online learning component that can help bridge this gap by learning through interaction. 

%3. Online learning differentiator and GPs: 
Most work on online learning focuses on estimating particular model parameters or uncertainty pertinent to the robot model itself rather than interactions with objects. 
One popular approach is the use of Bayesian non-parametric models for online learning \cite{Orbanz2010}. 
% These models generate predictions using the observed data.
These can readily incorporate priors, these can be in the form of offline-learned base models, which enable predictions in low-data scenarios.
They provide interpretable confidence metrics around their predictions in the form of a posterior distribution. 
These methods are data-driven and create Bayesian models on an effectively infinite-dimensional parameter space where the complexity of the model is allowed to grow with the size of the data \cite{10.5555/1162254}. 
% The adaptive nature of these models enables them to make informative predictions with low amounts of data. They also capture new and more complex behaviors as more data becomes available.
In particular, in this work, we use a form of Bayesian modeling called Gaussian processes (GPs) to model the robot-object interaction attributes. 
% In this work, we utilize Bayesian non-parametric methods for online learning. 
% These methods provide Bayesian models on effectively infinite-dimensional parameter spaces where the complexity of the model is allowed to grow with the size of the data \cite{10.5555/1162254}. 
% The adaptive nature of these models enables them to make informative predictions with low amounts of data. They also capture new and more complex behaviors as more data becomes available.

\begin{figure}[t]
  \centering
  \includegraphics[width=\linewidth, trim={25mm 0 25mm -5mm}, clip]{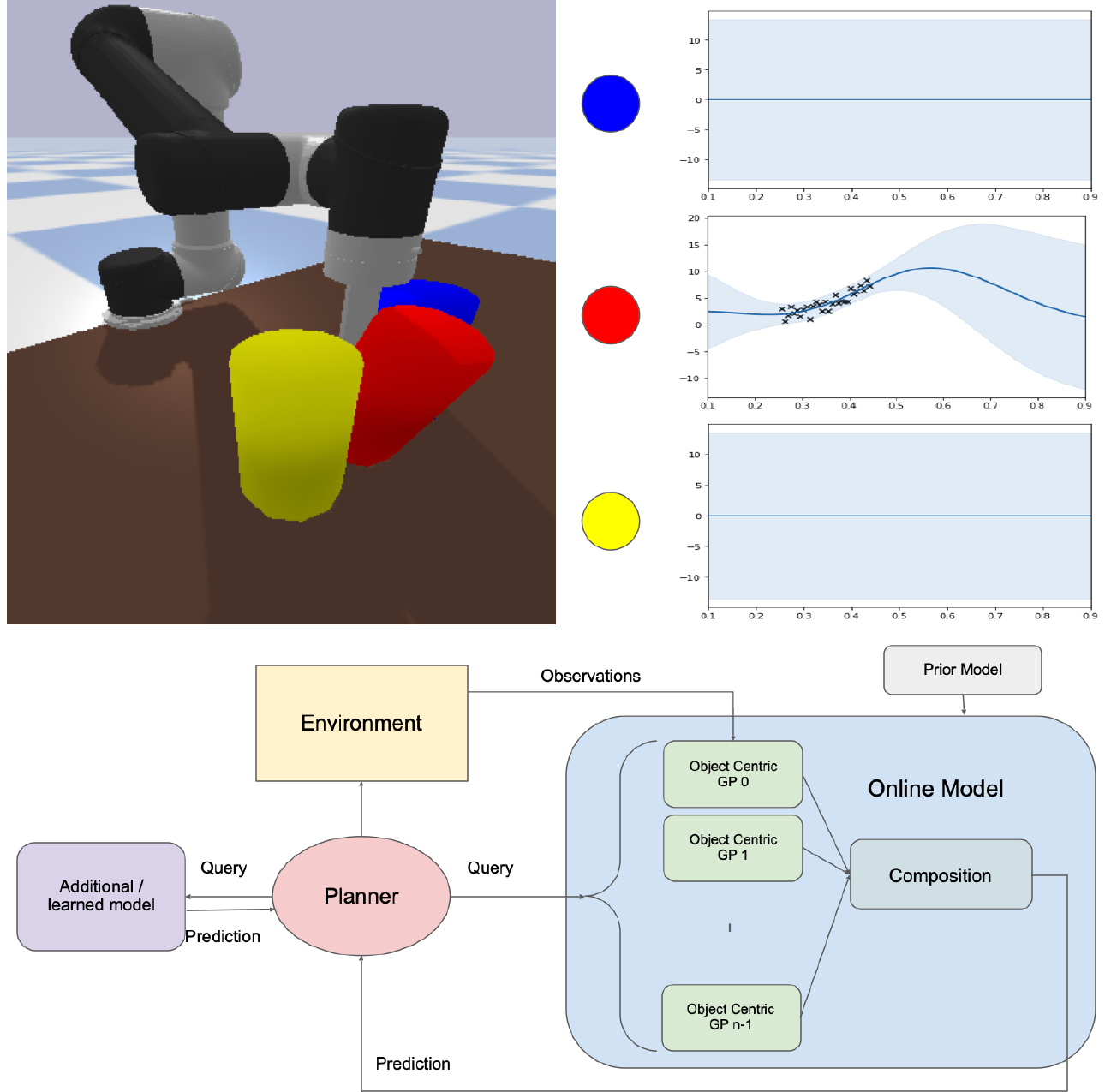}
  % \includesvg[width=\linewidth]{images/illustrative_image}
  \caption{Method Overview: Top: We apply our framework to the task of reaching a goal by pushing unknown objects on a table without knocking any over. 
  The robot has no prior knowledge of how to interact with the objects, and reasons online using composable object-centric Gaussian process regression. 
  A cartoon example of the robot's learned model is shown in the top figure. 
  Bottom: The chart shows an overview of our method and how our online model would fit into a general planning framework.\vspace{-2em}}
  \label{fig:front}
\end{figure}

% \begin{figure}[t]
%     \centering
%     \includegraphics[width=\linewidth]{example-image-a}
%     \caption{Caption}
%     \label{fig:my_label}
% \end{figure}
%4. We've established that we need online learning, now we need a better representation. 
%- Manipulation and online learning are both difficult tasks.

GPs are data-driven and rely on a new observation's similarity to the support set formed by its training data points. 
In interaction-based tasks, this can become challenging as the state of the objects and the robots, which form the datapoints, are constantly changing. 
The notion of finding the right data representation for a task has been very popular in offline learning methods \cite{yuan2021sornet}. 
% - \shnote{something about why online learning/GPs is still challenging when dealing with object interactions. Motivates finding a better representation for online learning to model interactions.}
% - \shnote{This notion of finding the right representation for a task has been very popular in offline learning methods}%One of the keys to the recent success of offline learning across a wide variety of different tasks has been utilizing the correct representation to learn the task at hand. 
Because the interaction between a robot and an object during manipulation tasks is largely object-centric in nature, recent literature in offline learning for manipulation have used object-centric representations to improve automation \cite{DBLP:journals/corr/abs-1907-03146}, \cite{yuan2021sornet}, \cite{zhu2022viola}. 
%This because the interaction between a robot and a object during these tasks is largely object-centric in nature. 

Inspired by these recent developments, we apply object-centric representations to improve \textit{online} learning for manipulation tasks. Our framework is shown in Fig.~\ref{fig:front}. We show that using an object-centric representation for online learning can be beneficial for capturing task-relevant features in our input representation and allow the model to learn better and be used online for the task at hand. 
%\nsnote{Sylvia Please look at this: }
%\shnote{reference framework in front figure, experiment on robot manipulator also shown in front figure.}

%In homes, the use of robotic manipulation is being explored for tasks including cooking, cleaning, and general household assistance. 
%These improvements in general household automation have the capability of revolutionizing the quality of life for the elderly and people with disabilities by granting them a degree of independence that they could not otherwise have. 
%One large barrier in realizing the goals of widespread automation of robotic manipulation is that it is quite difficult to perfectly capture and predict how a robot’s interactions will affect objects. 
%This is especially pertinent to the complex real world environments that can be seen in households and certain industrial applications. 

% - They provide models that: allow us to incorporate a prior, and compute a useable posterior that captures confidence in our predictions, they provide a very expressive model that can grow with 
% - able to be readily updated online 
% - prior - in low data scenarios 
% - posterior - usable confidence metric 
% - expressible and complexity grows with data seen and able to outgrow priors 
% - In particular we utilize Gaussian Processes 

\section{Related Works}
\label{related_works_section}
% \shnote{suggested ordering (once everything has been added):
% \begin{itemize}
%     \item Offline learning-based methods for interactive environments
%     \item Object-centric stuff (should end this part with Motivate learning online)
%     \item Online learning-based methods, Bayesian/GP stuff
%     \item Anything that has been done for online learning / interaction, sensing/estimating parameters of objects (should end this part with Motivating need for new representation for online learning / interaction)
% \end{itemize}}
% Previous active tracking.
Different sensor modalities using images \cite{9981966}, sound \cite{9830839}, and touch \cite{https://doi.org/10.1002/aisy.201900025} have been proposed in the literature to capture object-centric properties like deformation, relative pose, mass, friction, and texture. But these sensor values are subject to noise and need to be actively tracked. Numerous works have shown the benefits of tracking these errors for downstream robotic tasks \cite{9551655, 8263867, doi:10.1177/0278364911406562}. In \cite{10.5555/2908675.2908676}, the authors use GPs to estimate object deformation. The model uses prior data to fit the GP model and cannot generalize to new objects with different material characteristics. In \cite{7803326}, the authors actively track deformations using GPs, but it is not generalizable to other modalities. An application of this work looks at how environment uncertainty can be reduced by moving objects occluding the sensors \cite{7759112}. Still, it is not object-centric and doesn't consider how to interact with the environment. The authors in \cite{9561483} track the noise in pose estimates using an ensemble of learning models, but these cannot capture object properties like friction and mass. 

% Learning-based object representations:
Recent works have looked at object representation, all specifically using neural networks. Zhu et al. \cite{zhu2022viola} propose an object-centric learned representation using different camera views and proprioceptive data and uses the fused features to accomplish downstream manipulation tasks. Similarly, SORNet \cite{yuan2021sornet} uses a transformer-based architecture to generate latent embeddings for different objects that generalize to objects with similar shapes and textures. Still, since it uses images, it can't capture physical properties like friction and mass. Kofinas et al. \cite{kofinas2021rototranslated} propose object representation that is rotation and translation invariant, which makes learning more efficient. %Self-supervision can be used to learn the underlying object representations from videos of robot interaction with the scene\cite{https://doi.org/10.48550/arxiv.2205.06333}. The learned representation improves performance from downstream tasks such as behavior cloning and object localization. 

Most similar to our work with regards to learnable representations, the authors in \cite{9808141} describe an object-centric embedding specifically for task and motion planning. The authors train an encoder-decoder structure by optimizing for task representations and pixel-wise segmentation of images. These models require large datasets for training, and the generated latent representations are difficult to interpret. In unstructured environments, we are data deprived, and data is costly to acquire, rendering these neural network representations unrealistic to train and inadapatable. %On the other hand, non-parametric methods like GP require fewer data points and can model complex functions online.

% 2. Introduction to non-parametric methods
% \shnote{same this paragraph. the preliminaries/background section should just lay out the necessary concepts/math required to understand the new work. All of the motivation should be covered by intro/related work}
% In this work, we utilize Bayesian non-parametric methods for online learning. 
% These methods provide Bayesian models on effectively infinite-dimensional parameter spaces where the complexity of the model is allowed to grow with the size of the data \cite{10.5555/1162254}. 
% The adaptive nature of these models enables them to make informative predictions with low amounts of data. They also capture new and more complex behaviors as more data becomes available.

In summary, there continue to exist many challenging unstructured environments where interaction dynamics are unknown and must be estimated through online interaction. Most prior object representation work have focused on neural networks that are data-hungry to train and difficult to adapt online. On the other hand, methods that leverage online adaptation tend to be specialized towards tracking a singular or very small set of parameter errors in a given system model, with fewer considering the involved challenge of learning the potentially nonlinear, potentially stochastic, parameter function online.
%A large proportion of previous works avoid the issue entirely and focus on learning models in an offline setting. 
To address this gap, we described the paired use of (i) non-parametric methods that can learn, online, a statistical model of the measureable outcomes of interaction, and (ii) a task-relevant, object-centric representation that result in more scalable online learning. 

\section{Gaussian Process Regression}

GPs are Bayesian non-parametric models that capture the distribution over continuous functions using a set of Gaussian random variables. The distribution over all functions $f:\mathbb{R}^m\rightarrow\mathbb{R}$ is parameterized with a mean function, $\mu_p(x)$, and a covariance or kernel function, $k(x,z)$, written as
\begin{align}
    f(x) &\sim \mathcal{GP}(\mu_p(x), k(x,z)) \\
    \mu_p(x) &=\mathbb{E}[f(x)] \\
    k(x, z) &=\mathbb{E}[(f(x)-\mu_p(x))(f(z)-\mu_p(z))]
\end{align}
where $\mathbb{E}[\cdot]$ represents the expectation operation.

Given a set of training data inputs $X = \{x_{0}, x_{1} \dots x_{n}\}, x_{i} \in \mathbb{R}^{m}$ and their corresponding noisy target values $Y=\{y_1, y_2, \ldots, y_n\}$ where $y_i = f(x_i) + \bm{N},\bm{N}\sim \mathcal{N}(0, \sigma_n^2)$, the posterior mean and variance for a new point $x^*$ is given by:
\begin{align}
   & ~~~~~~~f(x^*) | X, Y, x^*  \sim \mathcal{N}(\mu^*, \sigma^*)\\
    \mu^* &= \mu_{p}(x^*) + K(X, x^*)^{T}K_y^{-1} (Y - \mu_p(X)) \label{eq:gp_mean} \\
    \sigma^* &= k(x^*, x^*) - K(x^*, X) K_y^{-1} K(X, x^*)\label{eq:gp_var}
    % \sigma(x) &= k^{*}(x, x) \label{eq:gp_var} 
\end{align}
where $K_y=K + \sigma_{n}^{2}I, K\in\mathbb{R}^{n\times n}$ and $K(X, x^*)^T=K(x^*, X)\in\mathbb{R}^{n}$. The matrix $K$ is constructed by comparing all pairs of points in the given dataset using the kernel function i.e., $K(i, j)=k(x_i, x_j)$. Similarly, the vector $K(X,x^*)$ is constructed by comparing all values in $X$ with $x^*$. The prior mean function, $\mu_{p}(x)$, can be set to 0 without loss of generality.

For this paper, we used the radial basis function (RBF) kernel function for $k$ given by
\begin{equation}
    k(x,z) = \alpha \exp\left(-\tfrac{1}{2}l^{-2}(x - z)^T(x-z)\right)
\end{equation}
where scaling factor $\alpha$ and lengthscale $l$ are hyperparameters that can be tuned. Any kernel function that belongs to the class of covariance functions can be chosen. An example of an alternate kernel is the forward kinematics kernel \cite{das2019forward}.

Due to the non-parametric nature of this method, this model is able to fit diverse data types. With enough data, the posterior distribution will overcome the model priors. 
% \textcolor{red}{[Safe IML citation].}
% - In GP Regression we can set our prior by setting $\mu(x)$ and $k(x,z)$. 
% - Given this prior and training data we can compute our posterior. 
% - This posterior turns out to be another GP with mean: GP Regression mean function and covariance: GP Regression covariance function. 
% - The prior mean function is often set to $0$ without loss of generality
% - The prior covariance, $k(x, z)$, can be chosen to define a similarity metric between data points $x$ and $z$. 
% - The mean function can often be left at $0$ it is useful, particularly in low data scenarios, to aptly pick the covariance function. 
% - Since this method captures the distribution over all continuous functions with the aforementioned parameterization, we usually do not need to worry about fitting the data. 
% - With enough data, the posterior can overcome poor priors. 

\section{Object-Centric Representations for Online Learning}
\begin{figure*}[ht]
  \centering
  % \includesvg[width=\linewidth]{images/single_object_twocol_PUSH-cropped}
  % \includesvg[width=\linewidth, height=87mm]
  % \includesvg[width=\linewidth, height=80mm]{images/single_object_tipandpush}
  % \includesvg[width=\linewidth, height=80mm]{images/single_object_tipandpush_1}
  \includegraphics[width=\linewidth, height=80mm]{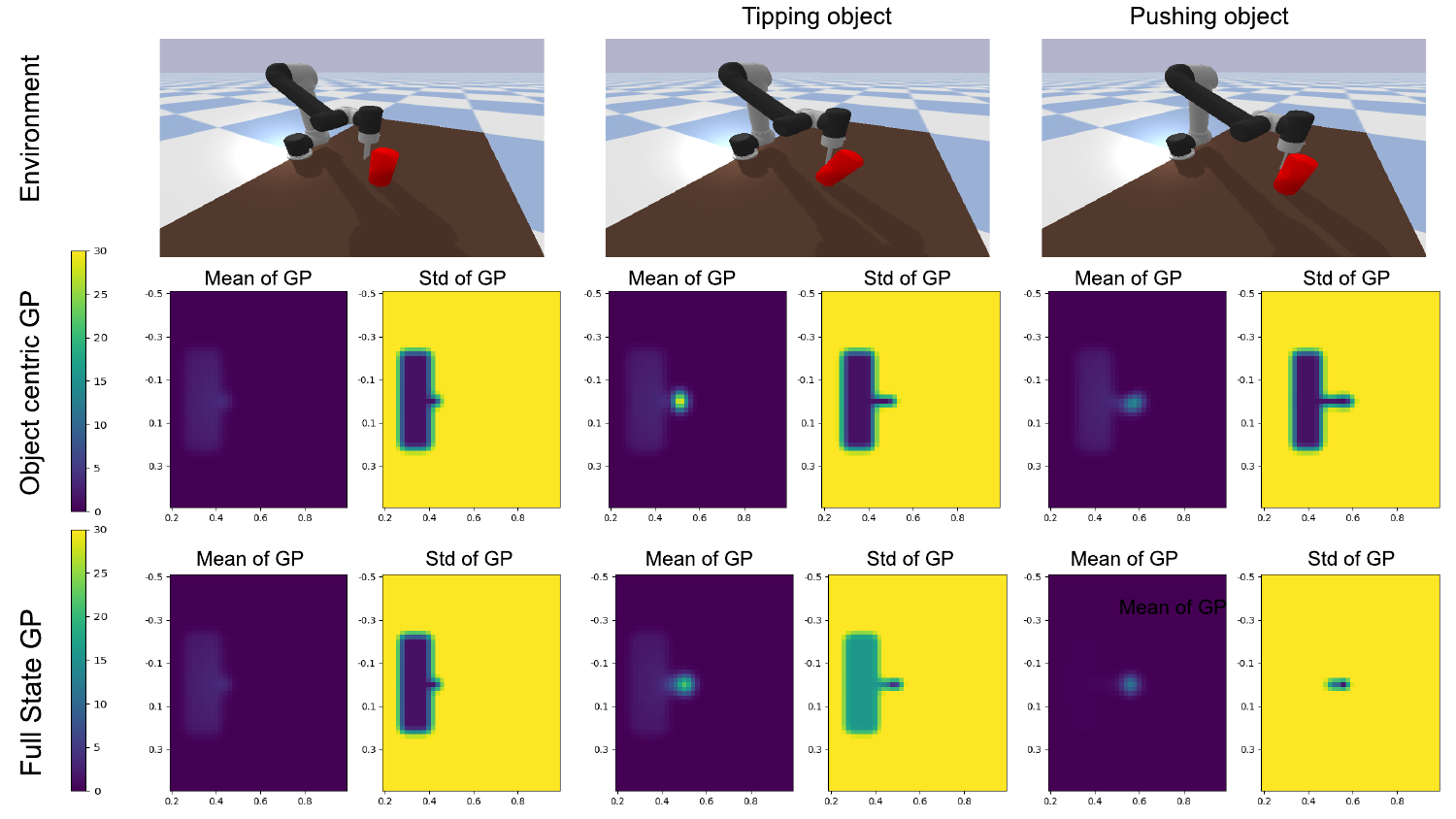}  
  \caption{Single object-centric GP: This figure demonstrates the benefit of using our object-centric GPs vs. a Full State GP baseline. 
  In this experiment, we are probing a single object. 
  We utilize the learned GP at different stages of interaction to predict the mean and variance of the tipping angle as a function of the robot state in the world frame and the current position of the object. 
  We show this in two setups, one where the interaction largely causes the object to tip: "Tipping object" and one where the interaction largely causes the object to translate "Pushing object". 
  Our object-centric representation allows our model to retain its predictive abilities despite changes in the object state, which cause the baseline GP quickly loses its ability to predict.
  %removing some white space with vspace
  \vspace{-2em} }
  \label{fig:single_object_GP}
\end{figure*}

% 1. What parameter we are estimating and where it comes from 
We model the attributes of the interaction dynamics between the robot and the objects in the environment using GPs by leveraging data collected online, purely through interactions. 
We need interaction in unstructured environments because the physical properties of these objects are not well-defined or may change with time. 
For example, consider an environment with 5 opaque bottles on a tabletop.
The robot would like to reach a goal without knocking over these bottles.
The properties of the bottles such as the center of mass and friction coefficients are challenging to model from visual sensors alone and are better understood by interactions.
In our experiments, we model the tipping angle of each object because it encapsulates different physical parameters of the objects.

% It is common to have a base model of the world for tasks that involve interacting with or manipulating objects. 
% This can be an offline learned model or one based on the physical intuition of the system. 
% However, it is often difficult to perfectly parameterize real-world systems in a manner that they can be precisely captured by these models. 
% Furthermore, for tasks involving interaction, some aspects of the environment may be impossible to capture perfectly without interacting with the environment online. 
% Thus, though we can rely on a base model to reason about the environment to some degree, we can use a learning method (e.g. GP Regression) to learn a predictive model for interaction while interacting with the environment online. 
% This model can directly estimate certain interaction attributes that would otherwise be unfeasible.

% For example, consider an environment with 5 opaque bottles on a tabletop. The robot would like to reach a goal without knocking over these bottles.
% The properties of the bottles such as the center of mass and friction coefficients may be hard to estimate from a distance.
% Without interacting with the bottles, it may be unfeasible to perfectly estimate how a bottle will tip when pushed.
% % This is as it is difficult to estimate parameters like mass, center of mass, friction coefficients etc. 
% Rather than attempting to estimate an unknown number of individual parameters that are coupled together, we can use a GP to directly estimate how the bottles will tip as a result of the robot interactions. 

% 2. Problem Statement 
Consider a simple illustrative running example shown in Fig.~\ref{fig:single_object_GP} with one object, $O_0$, and the robot. 
The robot has an end effector which is primarily responsible for interactions with the object. 
% \notemargin{For this paper we consider fairly homogeneous environments that do not have a large variation in their terrain for the bounds of the environment. }{
% technically can accommodate for variation in the terrain through something like another input parameter that was somehow estimated?}
For this paper, we consider fairly homogeneous environments that do not have a large variation in their terrain for the bounds of the environment. 
The object state is denoted by $O_0: p_{O_0} \in \mathbb{R}^3, q_{O_{0}}\in SO(3)$. 
Here $p_{O_0}$ and $q_{O_{0}}$ denotes the position and orientation of $O_0$ respectively. 
The robot end effector state is denoted by $A: p_{A}\in\mathbb{R}^3, q_{A} \in SO(3)$. 
Here $p_{A}$ and $q_{A}$ denote the position and orientation of the robot. All orientations are represented using quaternions in this paper.
These states are with respect to a world coordinate frame. 
% \shmargin{All these states: positions and orientations are with respect to a world coordinate frame.}{not a sentence}

% The interaction attribute that we are trying to model, for example the tipping angle of the bottle, is denoted by $y$. 
% \notemargin{The interaction attribute that we are trying to model is denoted by $y$. }{
% NOTE: is there a better way of presenting this explanation?}
The interaction attribute that we model is denoted by $y$.
This interaction attribute is a function of the object state in the world frame, the robot state in the world frame and the robot action in the world frame. 
For the purposes of this paper, we consider a quasi-static environment so that the interactions can be modeled using only the states without loss of generality. 
We can readily extend this to consider additional action parameters in our inputs and by transforming them with the same or similar object-centric representation. 
The true value of $y$ can be computed using $f(O_0, A, \theta)$ where $\theta$ denotes an unknown number of independent parameters that specify the real system. 
In practice, we only get noisy observations of the interaction attribute $y^{t}$ paired with noiseless observations of the object and robot state: $O_{0}^{t}, A^{t}, y^{t}$ at any timepoint. 

% 3. Typical input to the GP 
A naive implementation of the GP uses the object state and the robot state $\{ O_{0}^{t}, A^{t}\}$, or some subset of each of the states, as an input and outputs $y$ in attempt to model the true function $f$. 
GP regression is highly reliant on its similarity metric or kernel function and the observed data points in order to model behavior at a new unseen datapoint. 
As seen in Eq. \ref{eq:gp_mean}, \ref{eq:gp_var}, it uses the similarity metric to compare the new input to the support set of inputs contained in its training dataset. 
In tasks involving interacting with and manipulating objects, the object states will change as a result of the robot motion. 
Changes in the object state will cause all future inputs to the GP to appear to be further from the support set of training points that may not contain the object state in question. 
As a result, when the object state is altered by the robot, the GP Regression model's predictive capabilities will falter, resulting in prediction values close to the GP's prior with a very high variance indicating a low measure of confidence. 
We can see this behavior in Fig.~\ref{fig:single_object_GP}. 
As the robot interacts with the object and alters its state, the predictive variance on previously seen robot states increases drastically, and the predictive mean falls back to the GP prior. 

% 4. Framing object-centricness and how it is good and geared for object interaction/manipulation tasks. 
Despite the change in the object state in the world frame, the model should still be able to rely on its support set of datapoints from past interactions and allow reasoning about new interactions in this updated object state. 
This is because these manipulation/interaction-based tasks are fairly object-centric in nature, as discussed in section \ref{related_works_section}. 
The interaction between the robot and the object is a function of the state/action of the robot in the relative frame of the object. 
Thus despite the object state changing, the model should be able to continue to reason about certain robot-object interactions. 
We leverage this understanding by learning the interaction attribute using an object-centric GP. 
The object-centric GP uses the state of the robot in the object frame $A^{t}_{O_{0}}$, or some subset of the state, as an input and outputs a prediction on the interaction attribute $y$. 
% \textbf{NOTE: you are inconsistent with action or state, say at the beginning that you can have actions or states but for this paper we will use just states and then carry forward}

% 4. Poses and coordinate transforms to get things into object frame 
To compute $A^{t}_{O_{0}}$ we start by using $O_{0}$ to create a transformation matrix $T_{w}^{O_0}$ to convert coordinates in world frame to coordinates in the frame of the object $0_{0}$. 
To get $T_{w}^{O_0}$ we start with the pose $O_{0}$. 
We first compute the rotation matrix $R_{O_0}^{W}$ using the quaternion specifying the orientation of pose $O_{0}: q_{O_{0}} = [q_{0}, q_{1}, q_{2}, q_{3}]$. 
\begin{align} \label{eq:quat_to_rot}
\footnotesize
    R_{O_0}^{W} = 
    \begin{bmatrix}
        1 - 2q_{1}^{2} - 2q_{2}^{2} & 2q_{0}q_{1} - 2q_{3}q_{2} & 2q_{0}q_{2} + 2q_{3}q_{1} \\
        2q_{0}q_{1} + 2q_{3}q_{2} & 1 - 2q_{0}^{2} - 2q_{2}^2 & 2q_{1}q_{2} - 2q_{3}q_{0} \\
        2q_{0}q_{2} - 2q_{1}q_{3} & 2q_{1}q_{2} + 2q_{3}q_{0} & 1 - 2q_{0}^{2} - 2q_{1}^{2}
    \end{bmatrix}
\end{align}
This rotation matrix is used with the position of $O_0: p_{O_0}$ to compute the desired transformation matrix $T_{w}^{O_0}$ at time $t$. 
\begin{align}
    T^{W}_{O_{0}} = 
    \begin{bmatrix}
        R_{O_0}^{W} & p_{O_0} \\
        \Vec{0} & 1 
    \end{bmatrix}
    , \Vec{0} = [0, 0, 0] \\
    T_{w}^{O_0} =  [T^{W}_{O_{0}}]^{-1}
\end{align}

This transformation matrix is utilized to transform coordinates, such as the state of the robot, into the object frame. 
$A^{t}_{O_{0}}:  p_{A^{t}_{O_{0}}}\in \mathbb{R}^3, q_{A^{t}_{O_{0}}}\in SO(3)$. 
$R^{W}_{A^{t}}$ is the rotation matrix corresponding to the orientation $q_{A^{t}}$, computed similar to \ref{eq:quat_to_rot}. 
\begin{align}
    \begin{bmatrix}
         R^{W}_{A^{t}_{O_{0}}} & p_{A^{t}_{O_{0}}} \\
         \Vec{0} & 1 
    \end{bmatrix} =  T_{w}^{O_0} 
    \begin{bmatrix}
        R_{O_0}^{W} & p_{A^{t}}\\
        \Vec{0} & 1 
    \end{bmatrix}
    , \Vec{0} = [0, 0, 0]
\end{align}
$R^{W}_{A^{t}_{O_{0}}}$ is the rotation matrix describing the orientation of $A^t_{O_{0}}$ in the frame of $O_{0}$, which can be converted back to a quaternion $ q_{A^{t}_{O_{0}}}$.
% \textbf{NOTE: Go through math, say how now object pose baked in so only input is the object-centric representation}
These transformed representations are used with the object-centric GP. 
The transformation can be augmented or simplified by leveraging geometric symmetries of the object or other task-specific simplifications. 
% - If you know object geometrical symmetries, can leverage that for things like orientation invariance (if object spins wont change things) etc. 

% 5. What does this do: benefits
This object-centric representation consolidates the object and robot state and switches the input space of the GP to be more task-relevant, allowing better online modeling. 
Since the interaction attribute should only be a function of the robot state relative to the object state this representation provides a better space to compare new datapoints to the support set of datapoints collected online. 
Thus even if the object state has moved in the world frame, the model can leverage previous datapoints and reason about new robot states. 
By consolidating the two state spaces, we also reduce the dimensionality of the input space of the GP. 
We can see this behavior in Fig.~\ref{fig:single_object_GP}. 
As the robot interacts with the object and alters its state we still maintain a high confidence over previous states that have been sampled. 

%Multiple objects: 
In the case of multiple objects $\{ 0, 1, \dots n-1 \}$ we can model the pairwise interactions between the robot and each object using a separate object-centric GP for each object. 
This can be done when the robot accounts for a majority of what is measured in the interaction attribute. 
Each of these GPs can be updated intelligently, based on the proximity of the robot, to improve efficiency. 
The outputs of these GPs can be combined coherently, based on the task. 
This combined output can be used by a high-level planner to make intelligent decisions on how to move through the environment and interact with objects. 
One example of how multiple object-centric GPs can be composed is illustrated in section \ref{experiments_and_results_section}. 

% - can have object-centric GP modeling the attribute for each object
% - Assuming that the interaction effects are largely from the robot and not due to inter-object interactions
% - We combine the predictions of the individual models depending on the relevant task and use that for planning in more complicated multi-object environments. 

% \textbf{7. Conclusion sentence. ?}
% - Thus this method provides an efficient way to learn and model unknown attributes online through interaction 

\section{Experiments and Results}
%\shnote{I'd like to see this from an angle where it is obvious that the cup is tipping. Ideally so that the cup is sort of to the right of the robot so it matches with your GP plots. Then I think you might just need two columns: one where the robot has moved but hasn't touched the bottle yet (similar to your column 2 but hasn't quite reached the red cup) and one where the robot has interacted with the cup (your column 3). Can then maaaaybe condense this figure enough to not need it centered across the whole page}

% \shnote{I like the frames you captured for the multi-object figure.}
\begin{figure*}[ht]
  \centering
  % \includesvg[width=\linewidth, height=88mm]{images/multi_object4.svg}
  % \includesvg[width=\linewidth, height=82mm]{images/multi_object4.svg}
  % \includesvg[width=\linewidth, height=82mm]{images/multi_object5.svg}
  \includegraphics[width=\linewidth, height=82mm]{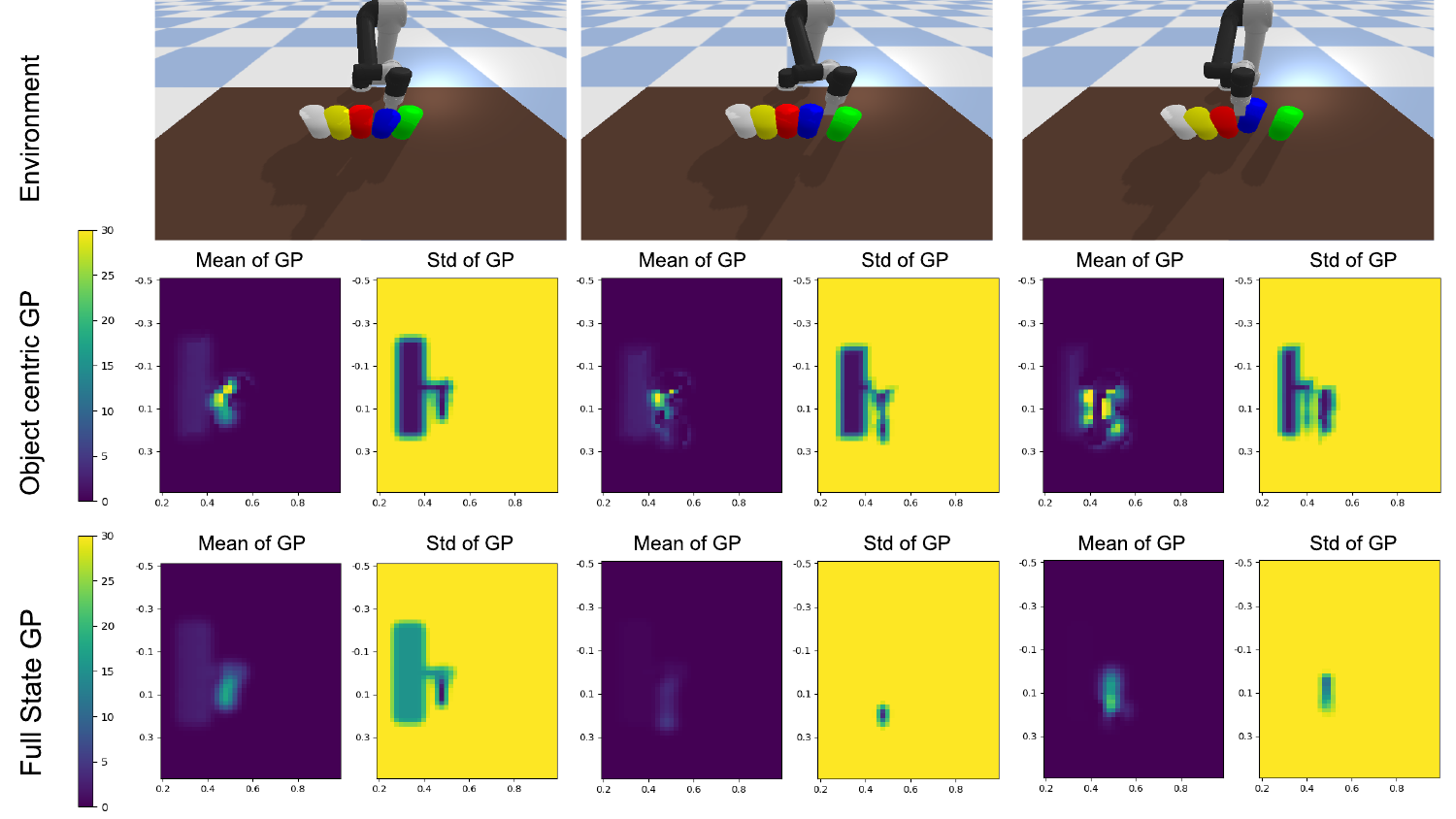}
  \caption{Multiple object-centric GPs: This figure demonstrates the benefit of using a composition of object-centric GPs to model robot-object interactions in environments with multiple objects. 
  The robot runs an open loop path and interacts with multiple objects. 
  We show frames of the robot as well as the mean and standard deviation of the tipping angle predicted by the online learned GP models as a function of the robot state in the world frame and the object states at that time point.
  Even in complex environments with multiple objects, our object-centric representation allows us to maintain a good model of the robot object interactions, while the full state GP fails once the objects have been interacted with. 
   %removing some white space with vspace
  \vspace{-2em}}
  \label{fig:multi_object_GP}
\end{figure*}

% \shnote{good structure for the flowchart. Could look prettier. Happy to help if need be.}

% 0. What problem we use to showcase method 
To showcase our method, we consider a problem with multiple objects on a tabletop with different attributes such as mass and center of mass. 
The interaction dynamics of the objects are unknown and estimated through online interaction. 
We consider a task involving non-prehensile manipulation where the robot must push the objects to reach a specified goal region. 
The robot wants to get to the goal without tipping the objects beyond a certain angle and knocking them over. 
We use GPs to directly learn the interaction by mapping the object-centric robot state to the tipping angle of the object, for each object. 
For these experiments the robot is constrained to motion in the $x,y$ plane at a fixed height, thus we only visualize the GP over robot states in the $x,y$ plane.  
We initialize our GPs with samples in the empty space around the robot for all experiments. 

% \textbf{NOTE: talk about constraining motion to x,y and that the states are just x,y coordinates.}

% \shmargin{\textbf{Single object open loop experiments: }}{if low on room, consider putting all of the single-object results in the methods section to combine with your running example.}
% In Fig.~\ref{fig:single_object_GP} we show the efficacy of our object-centric representation for the GP by demonstrating what our model learns when interacting with a single object. 
In Fig.~\ref{fig:single_object_GP} we demonstrate what our model learns when interacting with a single object. 
To do this, we run our robot in open-loop to probe a single object. 
We add the samples gathered during this open loop maneuver to update our GP model online. 
Each sample datapoint consists of the robot end effector state in the object-centric frame and the observed tip angle of the object. 
To visualize the GP at a point in time we plot the mean and standard deviation (std) of the predicted tipping angle for robot end effector positions in the world coordinate frame for the current object state. 
We compare this to the ``Full State GP" baseline. 
For this GP a sample datapoint's input contains the robot and object state in the world frame, with the corresponding object's tip angle as the output.

In Fig.~\ref{fig:single_object_GP}, we show the results of probing two different types of objects: A ``tipping object" whose interaction parameters cause it to mainly tip on contact, and a ``pushing object" whose parameters allow it to be pushed by the robot. 
As the object state changes, the full state GP fails to properly leverage support points from its training dataset to make valuable predictions about the tipping angle beyond the immediate position of the robot. 
This occurs when the object state changes slightly due to tipping and is much more exaggerated when the object is pushed. 
In contrast, our object-centric representation enables the model to continue to make meaningful predictions by leveraging its support set in spite of changes in the object state.

\textbf{Multiple object open loop experiment: }
We repeat the same experiment to show this methodology scaling to multiple objects, as shown in Fig.~\ref{fig:multi_object_GP} 
The Full State GP takes the state of the robot and objects in the world frame as its input and maps it to the maximum tip angle in the environment. 
With the object-centric, approach we split the problem up to consider pairwise interactions between the robot and each object separately. 
We learn each interaction online with a separate object-centric GP. 
These GPs can then be used to predict pairwise interactions, and their outputs can be combined for the task at hand. 

This scenario focuses on predicting the worst-case tipping angle in the environment in terms of the probabilistic upper bound of our understanding. 
We first predict the mean and variance of the tipping angle with each of our GPs: $\{ (\mu_{0}, \sigma_{0}), (\mu_{1}, \sigma_{1}), \dots, (\mu_{n-1}, \sigma_{n-1}) \}$. 
The upper bound $u(\mu, \sigma) = \mu + \beta \sigma$ is then computed for each output: $\{u_{0}, \dots u_{n-1}\}$. 
The mean and variance prediction corresponding to the highest upper bound: $(\mu_{i}, \sigma_{i}) , i = \arg \max_{i} u_{i}$, is then used. 
% \textbf{Maybe add equations for this here}
% To do this, at each point we compute the upper bound for each GP and then return the mean and variance values corresponding to the largest upper bound. 
% The corresponding full state GP takes the full state of the robot and objects in the world frame as its input and directly maps it to the maximum tip angle in the environment. 
% From Fig.~\ref{fig:multi_object_GP} we can see that the object-centric representations enable useful predictions even after multiple objects moved while using the full state makes it challenging to make useful predictions after corresponding objects have begun to move from their original states. 
From Fig.~\ref{fig:multi_object_GP} we can see that the object-centric representations enable useful predictions, in stark contrast to the full state, even after multiple objects have begun to move from their original states. 

\begin{figure*}[ht]
  \centering
  % \includesvg[width=\linewidth, height=90mm]{images/plan3.svg}
  % \includesvg[width=\linewidth, height=75mm]{images/plan3.svg}
  % \includesvg[width=\linewidth, height=78mm]{images/plan4.svg}
  \includegraphics[width=\linewidth, height=78mm]{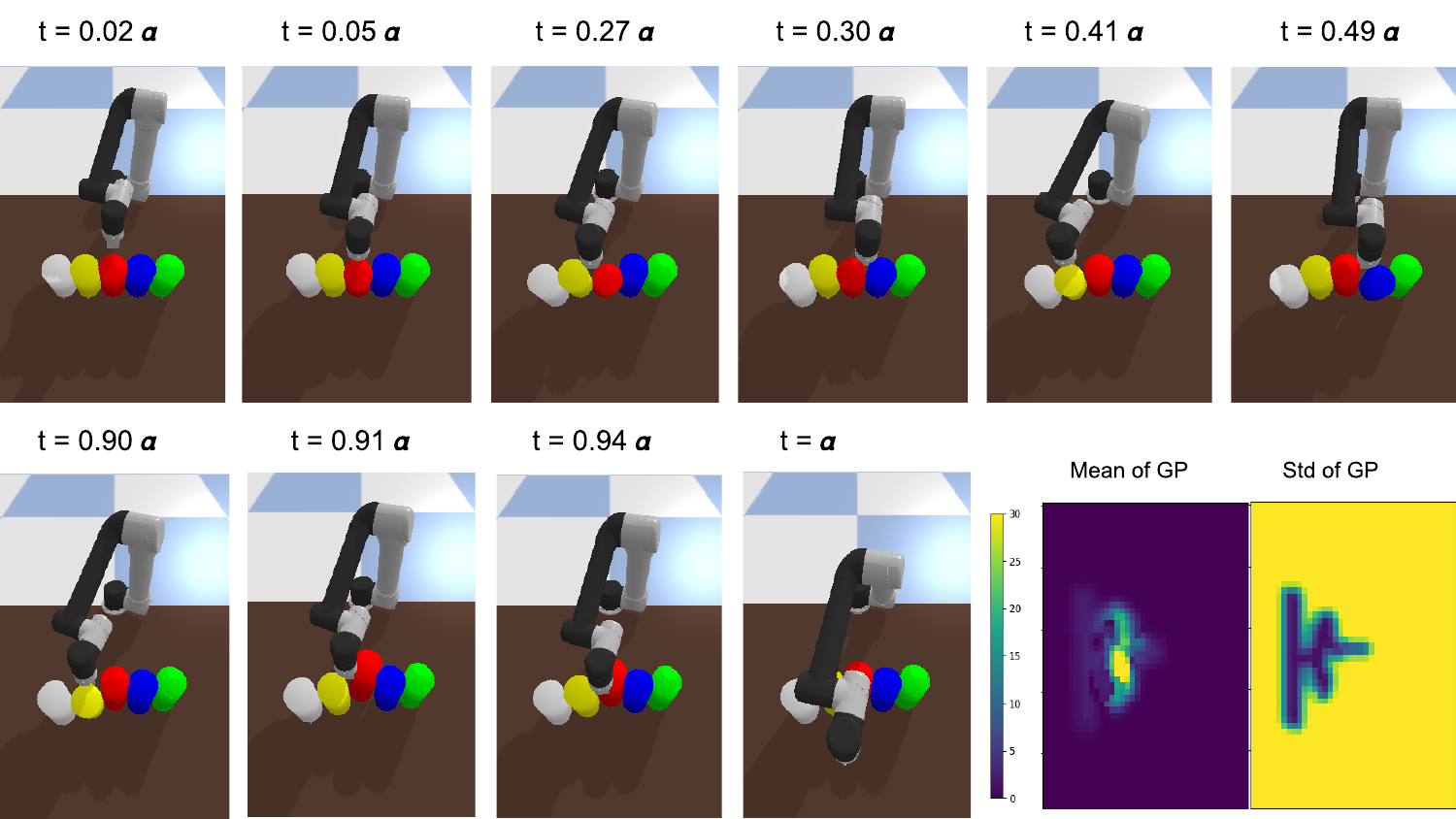}
  \caption{Online learning with object-centric GPs for Planning:  
  The robot is trying to get to a goal on the other side of several objects with unknown interaction dynamics without knocking any of them over. 
  The robot uses a composition of object-centric GPs to learn about its effect on the objects through online interaction. 
  The robot's planner queries the GP models and uses their predictions and confidence bounds to balance exploring the environment and exploiting what it has learned to get to the goal, the other side of the table. 
  We show the plan at different frames between timesteps $[0, \alpha]$. 
  We also show the mean and variance predictions of the worst-case tipping angle generated with our method at the final timestep. 
  This prediction is done over the robot states in the world frame with respect to the object states at the final timestep. 
   %removing some white space with vspace
  \vspace{-2em}}
  \label{fig:plan}
\end{figure*}

\textbf{Object-centric model integration with planner: } 
Our online learned models can be integrated with planners, as shown in Fig.~\ref{fig:plan}. 
The environment contains $5$ objects with unknown centers of mass, mass, and completely unknown interaction dynamics. 
The robot is attempting to get to a goal region at the other side of the table without knocking over the objects. 
The robot state space is bounded so that it can not trivially go around the objects. 
If the robot attempts to naively push through the objects, some objects will tip excessively and fall over. 
% - We showcase our methodology by integrating our model with a planner.  
To showcase our models, the planner lacks any prior on how the objects will move in response to interactions. 
Additionally, to showcase the learning capabilities of our GPs, they are initialized with a naive prior mean function, $\mu_{p}(x) = 0$, that indicates that the robot can move without affecting the objects. 
% The GPs are initialized with a set of robot positions sampled in a known empty region of the environment. 
% The planner interfaces with the online learned models to ensure that any positions that the robot is going to move through will not tip over the bottles so much that they would fall over. 
The planner uses the online learned models to ensure that the robot will not cause the bottles to fall over. 
The predicted mean and variance of the tipping angles are used, by the planner, to determine where the robot should sample to learn more about the environment, while balancing exploration and exploitation to get to the goal. 
% Experiment with object-centric model and planner: 
%The environment that we plan in is shown in the top left frame of Fig.~\ref{fig:plan}. 
% \textbf{Include this figure!} This is shown in figure [REFERENCE FIGURE] where a naive baseline planner attempts to get to the goal while completely disregarding the objects while planning. 
% Using our algorithm the robot is able to successfully learn about the environment through multiple sampling maneuvers without causing the objects to fall over. 
Using our algorithm the robot is able to successfully learn about the environment through multiple sampling maneuvers. 
The robot is able to leverage small gaps created between the objects as they are manipulated to squeeze between the objects in a manner that doesn't tip them over beyond a set threshold, to get to the goal region. 
The sampling maneuvers can be seen between timesteps $[0.04 \alpha, 0.9 \alpha]$ in Fig.~\ref{fig:plan}, before the robot exploits the gap between the bottles between timesteps $[0.9 \alpha, 0.94 \alpha]$

% 5. Maybe Experiment 2 with objects and planner
% - \textbf{Maybe show this: } We additionally show these results on a more complex environment containing $10$ unknown objects that the robot must interact with on its way to the goal at the end of the table. 

\label{experiments_and_results_section}

\section{Conclusion}

% 1. Re-iterate our method and the benefits it shows 
% - better represent task elevant features in our input representation
% - enables better online learning in unknown environments
% - we demonstrate the benefits of our methods and show its potential by integrating it with a planner to navigate through complex obstacle filled environments. 

% 1. Re-iterate our method and the benefits it shows 
% We demonstrate that through our use of task-relevant object-centric representations, we enable better online learning for robot-object interaction tasks in unknown environments. 
% We showcase the potential of our method by integrating it with a planner to navigate through complex obstacle-filled environments. 

We enable better online learning for robot-object interaction tasks in unknown environments through our use of task-relevant object-centric representations. 
We showcase the potential of our method by integrating it with a planner to navigate through complex obstacle-filled environments.

This work’s restriction to learning online limited our choice of predictive models for learning. 
Though GPs provide an efficient, non-parametric way to learn online they have limitations. 
These limitations include cubic computational complexity with increased samples, difficulty predicting in high dimensional spaces and difficulty in choosing hyperparameters. 
The drawbacks posed by these limitations can often be mitigated. 
Sparse GPs, local GPs or techniques such as thresholding the number of used predictive samples can help reduce predictive computational complexity. 
Creating an efficient kernel to leverage geometric similarities in the scene or focus on more relevant features can help extend this method to more complex high dimensional spaces. 
While certain hyperparameters can be set by maximizing the likelihood of the observed data, setting these parameters can also serve as a way to enforce principled priors on extending the learned model to uncertain, unsampled regions of the state space.

This work provides an initial step in interactive online learning to improve interaction-based tasks in unknown environments. 
%We seek to extend this work by pairing it with learned models for an improved prior understanding of the environment. 
%With improved priors and the use of more complex planners we seek to demonstrate this method on a variety of difficult household and industrial tasks. 
% Though we only discuss rigid objects in this work, object-centric representations can readily generalize to deformable objects. 
Such a object-centric representation can be used to incorporate richer state information that can ultimately lead to more intelligent interactions with deformable objects, for tasks such as tissue manipulation for surgical automation. 

\bibliographystyle{IEEEtran}
\bibliography{root}

\end{document}